%% file: 0-main.tex
\documentclass{bmvc2k}

\title{Human-Object Interaction Detection \textit{via} Weak Supervision}

\addauthor{Mert Kilickaya, }{kilickayamert@gmail.com}{1}
\addauthor{Arnold W.M. Smeulders}{arnoldsmeulders@uva.nl}{1}

\addinstitution{
 Visual Sensing Lab\\
 University of Amsterdam\\
 Amsterdam, Netherlands
}

\input{style}

\runninghead{Kilickaya, Smeulders}{HO-I Detection w/ Weak Supervision}

\def\eg{\emph{e.g}\bmvaOneDot}

\def\etal{\emph{et al}\bmvaOneDot}

\begin{document}

\maketitle

\begin{abstract}

The goal of this paper is Human-object Interaction (HO-I) detection. HO-I detection aims to find interacting human-objects regions and classify their interaction from an image. Researchers obtain significant improvement in recent years by relying on strong HO-I alignment supervision from~\citep{hicodet}. HO-I alignment supervision pairs humans with their interacted objects, and then aligns human-object pair(s) with their interaction categories. Since collecting such annotation is expensive, in this paper, we propose to detect HO-I without alignment supervision. We instead rely on image-level supervision that only enumerates existing interactions within the image without pointing where they happen. Our paper makes three contributions: \textit{i)} We propose Align-Former, a visual-transformer based CNN that can detect HO-I with only image-level supervision. \textit{ii)} Align-Former is equipped with HO-I align layer, that can learn to select appropriate targets to allow detector supervision. \textit{iii)} We evaluate Align-Former on HICO-DET~\citep{hicodet} and V-COCO~\citep{vcoco}, and show that Align-Former outperforms existing image-level supervised HO-I detectors by a large margin ($\textbf{4.71}\%$ mAP improvement from $16.14\%\rightarrow20.85\%$ on HICO-DET~\citep{hicodet}). 


\end{abstract}

\vspace{-4 mm}
\input{1-introduction}

\input{2-relwork}

\input{3-method}

\input{4-experiments}

\input{5-conclusion}

\bibliography{egbib}
\end{document}

%% file: style.tex
\usepackage{times}
\usepackage{epsfig}
\usepackage{graphicx}
\usepackage{amsmath}
\usepackage{amssymb}

\usepackage{bbm}

\usepackage{amssymb}
\usepackage{caption}
\usepackage[nocompress]{cite}
\usepackage{slashbox}
\usepackage[dvipsnames,table]{xcolor} 
\usepackage{tabularx,ragged2e}
\usepackage{spreadtab}
\usepackage{adjustbox}
\usepackage{booktabs}
\usepackage{xcolor}
\usepackage{multirow}
\usepackage{rotating}
\usepackage{color, colortbl}

\usepackage[inline]{enumitem}
\usepackage{color,soul}

\graphicspath{{./figures/}}

\DeclareGraphicsExtensions{.jpeg,.png,.eps,.pdf}

\definecolor{Gray}{gray}{0.9}

\definecolor{Gray}{gray}{0.85}
\definecolor{LightCyan}{rgb}{0.88,1,1}

\newcolumntype{a}{>{\columncolor{Gray}}c}
\newcolumntype{b}{>{\columncolor{white}}c}

\newcommand{\partitle}[1]{\bigbreak\noindent\textbf{#1}}

\makeatletter
\newcommand*{\rom}[1]{\expandafter\@slowromancap\romannumeral #1@}
\makeatother

\usepackage{mathtools}

\usepackage{pifont}
\newcommand{\cmark}{\ding{51}}%
\newcommand{\xmark}{\ding{55}}%

\newcommand{\Ree}{\mathbb{R}}

\makeatletter
\newcommand*\bigcdot{\mathpalette\bigcdot@{.5}}
\newcommand*\bigcdot@[2]{\mathbin{\vcenter{\hbox{\scalebox{#2}{$\m@th#1\bullet$}}}}}
\makeatother

\definecolor{applegreen}{rgb}{0.55,0.71,0.0}
\definecolor{babyblue}{rgb}{0.54,0.81,0.94}
\definecolor{azure}{rgb}{0.0,0.5,1.0}
\definecolor{budgreen}{rgb}{0.48,0.71,0.38}
\definecolor{amaranthpurple}{rgb}{0.67,0.15,0.31}

\newcommand{\etal}{\textit{et al}. }
\newcommand{\ie}{\textit{i}.\textit{e}., }
\newcommand{\eg}{\textit{e}.\textit{g}. }

\usepackage{cleveref}

\crefformat{section}{\S#2#1#3} 
\crefformat{subsection}{\S#2#1#3}
\crefformat{subsubsection}{\S#2#1#3}

\usepackage{booktabs,arydshln}

\makeatletter
\def\adl@drawiv#1#2#3{%
        \hskip.5\tabcolsep
        \xleaders#3{#2.5\@tempdimb #1{1}#2.5\@tempdimb}%
                #2\z@ plus1fil minus1fil\relax
        \hskip.5\tabcolsep}
\newcommand{\cdashlinelr}[1]{%
  \noalign{\vskip\aboverulesep
           \global\let\@dashdrawstore\adl@draw
           \global\let\adl@draw\adl@drawiv}
  \cdashline{#1}
  \noalign{\global\let\adl@draw\@dashdrawstore
           \vskip\belowrulesep}}
\makeatother

\definecolor{LightCyan}{rgb}{0.88,1,1}

%% file: 1-introduction.tex
\section{Introduction}

This paper strives for Human-object Interaction (HO-I) detection from an image. HO-I detection receives an astounding attention from the community recently~\citep{hicodet,gkioxari2018detecting,gao2020drg,ican,gupta2018no,hou2020visual,kim2020detecting,liao2020ppdm,liu2020amplifying,tamura2021qpic,chen2021reformulating,kilickaya2020self,kilickaya2021structured}, thanks to the large-scale benchmark of HICO-DET~\citep{hicodet}. The goal is to identify the tuples of \texttt{<human, object, verb, noun>} from the input, where human-object is an interacting bounding box pair, and verb-noun is the interaction type, such as ride-horse. 



\begin{figure}[t]
    \centering
\includegraphics[width=\textwidth]{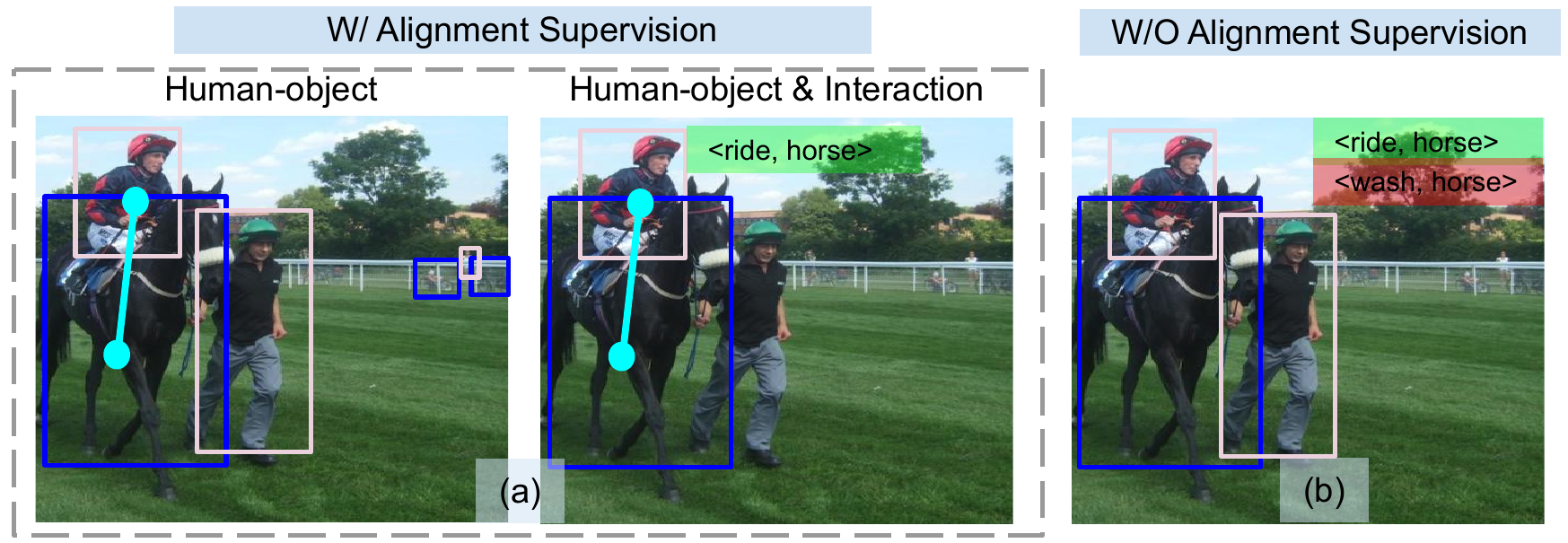}
 
 
 \caption{Alignment (left) \textit{vs.} Image-level HO-I supervision (right). \textit{a)} Alignment supervision annotates each human-objects, aligns humans to their interacting objects, then aligns human-objects to their type of interaction. \textit{b)} Image-level supervision only lists existing interactions without pointing where they happen. Our goal is to detect HO-I \textit{without} costly alignment supervision, by only using image-level labels.}
    \label{fig:teaser}
\end{figure}

To tackle this problem, researchers leverage strong HO-I alignment supervision, see Figure~\ref{fig:teaser}-(\textcolor{red}{a}). Annotators first draw a bounding box around all humans and objects, then align humans with the object-of-interaction (\eg, rider and horse). Finally, they align the interaction category with each human-object pairs.

However, collecting such annotation is costly~\footnote{Try-it-yourself! 
\href{http://www-personal.umich.edu/~ywchao/hico/hoi-det-ui/demo_20171121.html}{HICO-DET-Annotator}}. Annotation costs time, since in a typical image there are tens of potential human-object interactors, if not hundreds. One can instead rely on image-level HO-I annotations, see Figure~\ref{fig:teaser}-(\textcolor{red}{b}). Image-level annotations enumerate existing HO-I within the image, without specifying where they occur. Image-level annotations are much faster and cheaper to collect.




There are few attempts to perform HO-I detection via image-level supervision~\citep{weaklyhoi1,weaklyhoi2}. Initially, Zhang~\etal~\citep{weaklyhoi1} proposes a two-stream architecture based on Region-FCN~\citep{rfcn}, focusing on the regional appearance of subject-objects and spatial relations. Later, Kumaraswamy~\etal~\citep{weaklyhoi2} adapted this technique for HO-I detection, and improve it via an additional stream of human pose. These techniques yield remarkable results on HICO-DET benchmark~\citep{hicodet} in the absence of alignment supervision. However, they are limited in three major ways: \textit{i)} These methods isolate human-objects from their context via Region-of-Interest (RoI) pooling~\citep{fastrcnn,faster-rcnn}, however, contextual information is crucial in understanding the interaction, \textit{ii)} The authors propose multiple streams of context to circumvent the missing contextual information, which increases model complexity. Increased model complexity results in low performance on especially rarely represented HO-I (\ie \texttt{<ride, cow>}) as we will show.  \textit{iii)} Hand-crafted context (\ie body-pose configuration using key-points) may not be sufficient to account for the complexity of HO-I detection problem. 

To that end, in this paper, we propose Align-Former, a visual-transformer-based architecture based on~\citep{detr}. Align-Former is a single-stream HO-I detector that is trained end-to-end using image-level supervision only. Align-Former is equipped with a novel HO-I Align layer that learns to align a few candidate target HO-I with predictions, allowing detector supervision. The decision of alignment is based on geometric and visual priors that are crucial in HO-I detection. 



This paper makes the following contributions: 

\begin{enumerate}[label=\Roman*.]

\item We propose Align-Former, an end-to-end HO-I detector that is supervised via image-level annotation. 

\item We equip Align-Former with a novel HO-I align layer, that learns to match few HO-I predictions with HO-I target(s), therefore allowing detector supervision. 
\item We evaluate Align-Former on HICO-DET~\citep{hicodet} and V-COCO~\citep{vcoco}, and show that Align-Former outperforms competing baselines with the same level of supervision (by $\textbf{4.71}$ mAP) on the large-scale benchmark of HICO-DET~\citep{hicodet}, especially within the low-data regime of rare categories (by $\textbf{6.17}$ mAP). 

\end{enumerate}

%% file: 2-relwork.tex
 \begin{figure}[t]
    \centering
\includegraphics[width=\textwidth]{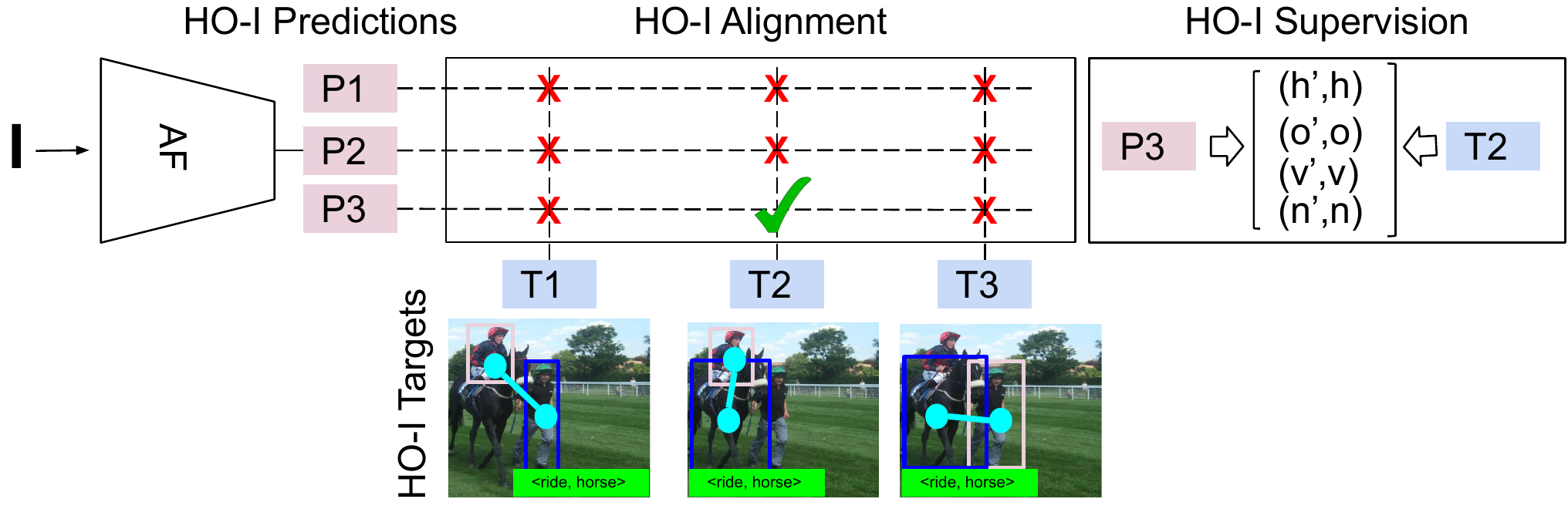}

  \caption{To perform HO-I detection via image-level supervision: \textit{i)} Align-Former maps the input image $I$ to HO-I predictions $P$ . \textit{ii)} We also prepare a set of HO-I targets by exhaustively matching human-object detections and list of interactions. \textit{iii)} Finally, we find the least costly prediction-target pair(s) (\ie $(T_2, P_3)$)  which will be used for detector supervision.}
  
    \label{fig:coreidea}
\end{figure}

\section{Related Work}

\partitle{Alignment-Supervised HO-I Detection.} In HO-I detection, the goal is to find quadruplets of \texttt{<human,object,verb, noun>} where human-object are bounding boxes and verb-noun are interaction pairs like \texttt{<ride, horse>}. Initially, HICO-DET authors collect more than $150k$ instance annotations to match humans to their interacted object, as well as to their interaction categories. Then, there has been a surge in detecting HO-I, initially via two-stage techniques~\citep{hicodet,gkioxari2018detecting,ican,gupta2018no,hou2020visual,liu2020amplifying}, and later by one-stage architectures~\citep{gao2020drg,kim2020detecting,liao2020ppdm,tamura2021qpic,chen2021reformulating} leveraging costly strong alignment supervision, see Figure~\ref{fig:teaser}-(\textcolor{red}{a}). 

In this work, our goal is to train HO-I detectors without alignment supervision, by only relying on image-level HO-I annotations. 

\partitle{HO-I Detection via Image-level Supervision.} Few works attempt to train HO-I detectors by only image-level supervision~\citep{weaklyhoi1,weaklyhoi2}. Initially, Zhang~\etal~\citep{weaklyhoi1} proposes a two-stream architecture based on Region-FCN~\citep{rfcn} to model the subject-object region appearance and spatial relations. Later, Kumaraswamy~\etal~\citep{weaklyhoi2} extends this approach via additional pose-stream. These methods operate on the isolated appearance of human-objects, neglecting the crucial context. Consequently, they supplement Region-FCN with additional streams, increasing the model size, decreasing the performance. 


To circumvent this, in this work, we propose a single-stream HO-I detector based on visual-transformer~\citep{detr}. Our network naturally encodes the surrounding context of human-objects thanks to self-attention~\citep{vaswani2017attention} and learns to align few candidate HO-I targets with HO-I predictions to perform detector supervision, see Figure~\ref{fig:coreidea}. 


\partitle{Discrete Variable Sampling in Computer Vision.} In this work, we treat HO-I target alignment as a hard-valued, discrete variable sampling: Amongst all possible target-prediction pair(s), which subset(s) should be selected for detector supervision? Such decision is non-differentiable therefore ill-suited in convolutional network training. To that end, we resort to a continuous relaxation procedure named Gumbel-Softmax trick, which allows end-to-end training via discrete variables~\citep{jang2016categorical,maddison2016concrete}. Gumbel-Softmax has successfully been used to sample convolutional layers~\citep{featuremapgating}, filters~\citep{filtergating} or channels~\citep{channel}. 

In this work, we adapt Gumbel-Softmax to select the target HO-I for detector supervision.

%% file: 3-method.tex
 \begin{figure}[t]
    \centering
\includegraphics[width=\textwidth]{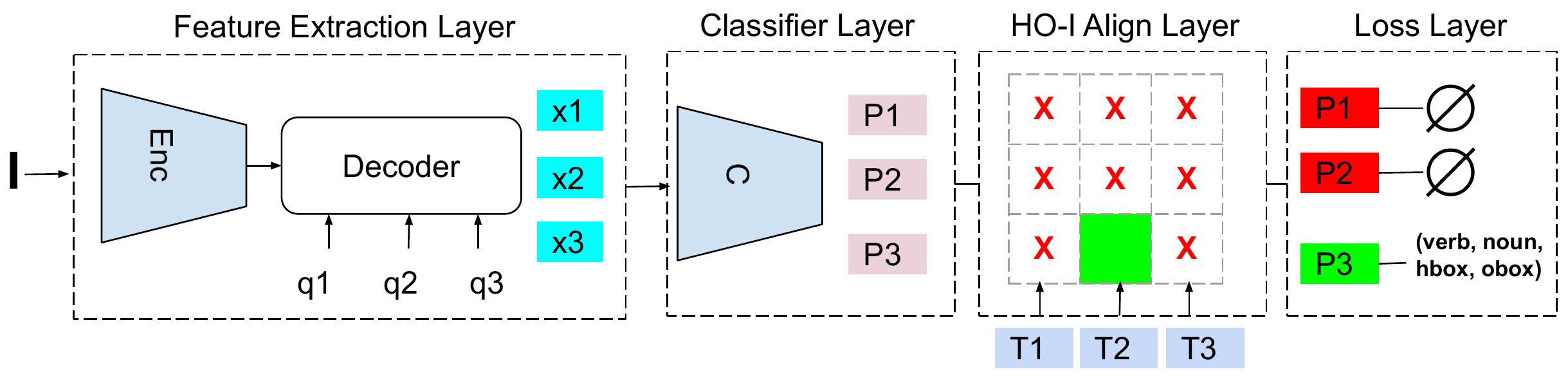}

\caption{Align-Former consists of four main layers. \textbf{Feature Extraction Layer} is an Encoder-Decoder-based visual-transformer that extracts a set of human-object features $x_i$ using the positional queries $q_i$. Then, \textbf{Classifier Layer} generates HO-I predictions $P$ in the form of human-object bounding boxes and verb-noun classes. \textbf{HO-I Align Layer} compares HO-I predictions $P$ with potential HO-I targets $T$ to find few-matching pair(s) that are used for HO-I detector supervision using \textbf{Loss Layer}.}

    \label{fig:method}
\end{figure}

\section{Align-Former for HO-I Detection}

\partitle{Method Overview.} An overview of our technique is presented in Figure~\ref{fig:coreidea}-\ref{fig:method}. The goal of our network $g_{\theta}(\cdot)$ is to produce HO-I prediction tuples given an image $I$ as $I\xrightarrow{g_{\theta}(\cdot)}t^{\prime}$. Here, HO-I prediction is of size $P$ and represented via $t^{\prime} = (h^{\prime},o^{\prime},v^{\prime},n^{\prime})$, where $(h^{\prime}\in\Ree^{P\times4} ,o^{\prime}\in\Ree^{P\times4})$ are human-object bounding box predictions, and $(v^{\prime} \in\Ree^{P\times V},n^{\prime} \in\Ree^{P\times N})$ are verb-noun class predictions for $V$ verbs and $N$ nouns.

Then, assume we have access to a set of HO-I targets of size $T$ with the same structure $t = (h\in\Ree^{T\times4},o\in\Ree^{T\times4},v\in\Ree^{T\times V},n \in\Ree^{T\times N})$. To supervise Align-Former, we propose to minimize the following objective:
\begin{align}
    \min_{\theta}(A \times t, t')
\end{align}
\noindent where we omit $\theta$ from now on for clarity. $A$ is a binary matrix of size $P\times T$ where only few entries are non-zero. $A$ is applied separately on all tuple members, as $A \times t = (A\times h, A\times o, A\times v, A\times n)$. Here, $A(i, j) = 1$ means prediction $i$ matches (\ie aligns) with target $j$ to use in supervision. Similarly, $A(i, j) = 0$ indicates target $i$ should not be used in detector supervision. To identify which target-prediction pairs should be used in detector supervision, we rely on geometric and visual priors detailed later.  






Finally, replacing $t^{\prime}$ with $g(I) = C(Dec(Enc(CNN(I)), Q))$ yields:
\begin{align}
  \min (A \times t, C(Dec(Enc(CNN(I)), Q)))
\end{align}
\noindent which is detailed in four Sections: 


\begin{itemize}[leftmargin=*]
    \item \textbf{HO-I Align Layer (\cref{sec:hoialign})} generates the alignment matrix $A$ that pairs few HO-I prediction(s) with HO-I target(s),
    \item \textbf{Classification Layer (\cref{sec:hoiclass})} generates human-object bounding boxes and verb-noun classification via $C(x)$ using human-object features $x$,
    \item \textbf{Feature Extraction Layer  (\cref{sec:hoifeature})} generates features via $x = Dec(Enc(CNN(I)), Q)$ via positional queries $Q$ using Encoder-Decoder architecture,
    \item \textbf{HO-I Loss Layer (\cref{sec:loss})} computes the human-object box and verb-noun classification losses to supervise the detector with the generated HO-I targets $t$.
\end{itemize}

\subsection{HO-I Align Layer}
\label{sec:hoialign}

HO-I align layer consists of two sub-layers, \textit{i)} Prior layer that judges the compatibility between all HO-I targets and predictions, \textit{ii)} Discretization layer that binarizes the likelihood values to obtain the final hard-alignment. 

\subsubsection{Discretization Layer}


Assume we are given a scoring function $S\in \Ree^{P\times T}$ where $S(i, j)$ encodes how compatible HO-I prediction $t^{\prime}_{i}$ and HO-I target $t_{j}$ matches. Our goal is to discretize this matrix to obtain the final hard-valued alignment decision. 

To perform this, we discretize $S$ such that only few members will be non-zeros. Specifically, given raw values of $S$, we apply the following operation: 

\begin{align}
    A = \sigma(S + G)  \geq \delta
\end{align}

\noindent where $\delta = 0.5$ is the hard-threshold value, $G$ is the Gumbel noise~\citep{jang2016categorical,maddison2016concrete} added to the matrix $S$ for regularization, and $\sigma(\cdot)$ is the sigmoid activation to bound $S$ between $[0,1]$. Note that Gumbel-noise is crucial to avoid any degenerate solutions like all $1$s. 

This operation yields the binary alignment matrix $A \in \{0,1\}$ where only a few entries are non-zero. 


\subsubsection{Prior Layer}
\label{sec:priors}


To compute the compatibility between HO-I targets \& predictions, we resort to a convex combination of geometric and visual priors as $S = \alpha_{g} * GP + \alpha_{v} * VP$. Our intuition is that for an HO-I target to be a good candidate for detector supervision, it needs to be compatible both in terms of human-object bounding boxes (geometric) and verb-noun classes (visual). 


\partitle{Geometric Prior $GP(\cdot)$} computes the bounding box compatibility of human-objects via $L_{1}$ distance as: 

\begin{align}
GP = \exp(- \frac{\sum_{ij} \lVert h'_{i} - h_{j} \rVert + \lVert o'_{i} - o_{j} \rVert}{\tau})
\end{align}


\noindent where the exponential function $\exp(\cdot)$ converts the distance values to similarity where $\tau = 1$. 


\partitle{Visual Prior $VP(\cdot)$} computes how well a given target-prediction pair matches in terms of HO-I classes. Remember that our HO-I targets enumerate existing HO-I from the image in terms of verb-noun pairs. Therefore, $VP(\cdot)$ is calculated as: 


\begin{align}
    VP = v' * v^T + n' * n^T
\end{align}


\noindent where verb-predictions are of size $v' \in \Ree^{P\times V}$ and verb-targets are of size $v \in \Ree^{T\times V}$ for $V$ distinct verbs. Similarly, noun-predictions are of size $n' \in \Ree^{P\times N}$ and noun-targets $n' \in \Ree^{T\times N}$ for $N$ distinct nouns.

\subsection{HO-I Classification Layer}
\label{sec:hoiclass}



Classifier layer is responsible for generating HO-I predictions $t^{\prime}$ consisting of human-object bounding box predictions $(h^{\prime}, o^{\prime})$ as well as verb-noun category predictions $(v^{\prime}, n^{\prime})$. 


\partitle{Human-Object Bounding Box Classifiers} are two multi-layer perceptrons $g^{h}(\cdot)$ and $g^{o}(\cdot)$ that maps human-object features $x$ to coordinates as $(h^{\prime}, o^{\prime}) = (\sigma(g^{h}(x)), \sigma(g^{o}(x)))$.  

\partitle{Verb-Noun Classifiers} are also two multi-layer perceptrons as $g^{v}(\cdot)$ and $g^{n}(\cdot)$ that learns to map human-object features $x$ to corresponding verb-nouns as $(v^{\prime}, n^{\prime}) = (\sigma(g^{v}(x)), (g^{n}(x)))$.

\subsection{HO-I Feature Extraction Layer}
\label{sec:hoifeature}


Our backbone needs to encode: \textit{i)} Object-object relations, \textit{ii)} Relative object positions that are critical to perform HO-I alignment and detection. To that end, we implement the feature extractor as a visual-transformer based on DETR~\citep{detr}. The feature extractor yields human-object features $x \in\Ree^{P\times D}$, and consists of three sub-layers: Backbone, Encoder and Decoder, which are detailed below. 



\partitle{Backbone ($x = CNN(I)$).} Backbone is a deep CNN~\citep{resnet} that extracts global feature maps from the input image $I$ of size $x \in \Ree^{H\times W\times C}$ where $[H,W]$ are the height-width of the feature map, and $C$ is the number of channels. 



\partitle{Encoder ($x = Enc(x)$).} Encoder further processes the global feature map from the backbone to increase positional and contextual information. We first reduce the number of channels from the backbone to a much smaller size via $1\times1$ convolutions of $C\times D$. Then, the resulting feature map $\Ree^{H\times W\times D}$ is collapsed in the spatial dimension as $\Ree^{D\times HW}$ where each pixel becomes a "token" represented by $D$ dimensional features. Finally, this feature undergoes a few self-attention operations via few multi-layer perceptrons, residual operations, and dropout. At each step, pixel positions are added to the feature map to retain position information. 



\partitle{Decoder ($x = Dec(x, Q)$).} The Decoder is a combination of self-attention and cross-attention layers, which yields the final human-object features. The Decoder takes as input the Encoder output $x\in\Ree^{D\times HW}$ as well as fixed positional query embeddings $Q\in\Ree^{P\times D}$. Decoder alternates between the cross-attention between the feature map $x$ and $Q$, as well as self-attention across queries. Cross-attention extracts features from the global feature maps, whereas self-attention represents object-object relations necessary for HO-I detection. Decoder is implemented as multi-layer perceptrons. Final output is $x \in\Ree^{P\times D}$ that encodes positional and appearance-based representations of potential human-object pairs within the image. 


\subsection{HO-I Loss Layer}
\label{sec:loss}

Our loss function  ensures that the predicted human-object bounding boxes as well as the verb-noun predictions are in line with the aligned HO-I targets. 

The loss function $\mathcal{L}$ is a composite of bounding box, classification, and sparsity losses as $\mathcal{L} = \mathcal{L}_{box} + \mathcal{L}_{class} + \mathcal{L}_{sparse}$. Here, $\mathcal{L}_{box}$ computes the $L_1$ distances between human-object predictions and (aligned) targets as $\mathcal{L}_{box} = \mathcal{L}_{human} + \mathcal{L}_{object}$. And, $\mathcal{L}_{class} = \mathcal{L}_{verb} + \mathcal{L}_{noun}$ are implemented via classical cross-entropy. As there can be multiple verbs for each instance, we use sigmoid activation before computing the verb loss.  

\partitle{Sparsity Loss.} Finally, sparsity loss minimizes $\mathcal{L}_{sparse} = \frac{1}{P \times T}\sum_{ij} A_{ij}$ where $\frac{1}{P \times T}$ is a constant normalizing factor to bound the loss. This ensures the sum over all entries within the alignment matrix $A$ is minimized, leading to only few pairs of HO-I predictions and targets to be aligned for further supervision.  


\partitle{Implementation.} We set the number of predictions as $|P| = 100$. Our network is implemented using PyTorch~\citep{pytorch}. Feature size $D$ from the last layer of the Decoder is set to $D = 256$. Both human-object bounding box classifiers and verb and noun predictors are $2$-layer perceptrons with ReLU activation in between.Initial learning rate is set to $10^{-6}$ for the ResNet backbone and $10^{-5}$ for the rest of the parameters. We use weight-decay to regularize the network with $10^{-4}$. We train the network for $150$ epochs with an effective batch size of $16$ over $8$ GPU Titan cards. We decay the learning rate linearly with $10^-1$ after epoch $100$.  







%% file: 4-experiments.tex
\section{Experimental Setup}





\partitle{Datasets.} We experiment on two large-scale standard datasets, namely HICO-DET~\citep{hicodet} and V-COCO~\citep{vcoco}. \textit{i) HICO-DET} contains $38k$ images for training and $9.6k$ images for testing. Images contain $117$ distinct verbs and $80$ distinct nouns together, making $600$ \texttt{<verb, noun>} pairs. For each noun, there exists a case of "no-interaction", where at least a single human and the target object is visible, even though not interacting. We only use HO-I alignment annotations for testing, and not training, since our goal is to evaluate HO-I detection via image-level supervision. \textit{ii) V-COCO} builds upon MS-COCO~\citep{mscoco} where the authors annotate subset of images with human-object alignments and their (inter-)action. The type of interactions is riding, reading and smiling. The dataset exhibits $2.5k$ images for training, $2.8k$ images for validation, and $4.9k$ images for testing.

\partitle{Metric.} We use the mean Average Precision (mAP) metric for evaluation as is the standard~\citep{hicodet,vcoco}. A human-object interaction is true positive only if both humans and objects have an Intersection-over-Union with a ground-truth HO-I pair above $ > 0.50$ \textit{and} they are assigned to the correct interaction categories.

\partitle{Evaluation.} \textit{i) HICO-DET:} We use the evaluation code presented in the server~\citep{hicodetserver}.  We compute the mean over all three splits of full, rare, and non-rare in HICO-DET. We provide comparison on three standard splits. \textit{Full}: All $600$ categories, \textit{Rare}: $138$ categories with less than or equal to $10$ training instances, \textit{Non-Rare}: $462$ categories with more than $10$ training instances. \textit{ii) V-COCO:} We use the evaluation code presented in authors' code~\citep{vcocoserver}. We evaluate using three different standard scenarios. \textit{Agent}: We report the human interactor detection performance, \textit{Scenario-1}: We report the detection of humans and objects together, \textit{Scenario-2}: We report the detection of humans and objects where the object predictions for object-less interactions (\ie smiling) is ignored. 




\partitle{Baselines.} We compare Align-Former to \textit{i) Weakly-supervised HO-I detectors}:  PPR-FCN~\citep{weaklyhoi1} and MX-HOI~\citep{weaklyhoi2} that performs HO-I detection without alignment supervision. \textit{ii) Strongly-supervised variants}: To measure the upper bound performance as a reference, we also report MX-HOI and Align-Former performance via strong alignment supervision. 









%

\section{HO-I Detection on HICO-DET \& V-COCO}

\subsection{Comparison to The State-of-The-Art}


\begin{table}[h]
\begin{center}
\resizebox{\columnwidth}{!}{%
\begin{tabular}{llc|ccc}

\toprule 


Method   & Backbone & Alignment-Supervised? & Full & Rare & Non-Rare  \\ 
\midrule

PPR-FCN~\citep{weaklyhoi1} & ResNet-$101$  & \xmark & $15.14$ & $10.65$ & $16.48$  \\

MX-HOI~\citep{weaklyhoi2} &  ResNet-$101$  & \xmark & $16.14$ & $12.06$ & $17.50$  \\ 
\rowcolor{LightCyan}  Align-Former (ours)  & ResNet-$50$ & \xmark & $\underline{19.26}$ & $\underline{14.00}$ & $\underline{20.83}$  \\
\rowcolor{LightCyan} Align-Former (ours)  & ResNet-$101$ & \xmark & $\textbf{20.85}$ & $\textbf{18.23}$ & $\textbf{21.64}$  \\
  \cdashlinelr{1-6}
 MX-HOI~\citep{weaklyhoi2} &  ResNet-$101$  & \cmark & $17.82$ & $12.91$ & $19.17$  \\ 

 Align-Former (ours)  & ResNet-$50$ & \cmark & $25.10$ & $17.34$ & $27.42$  \\
 Align-Former (ours)  & ResNet-$101$ & \cmark & $27.22$ & $20.15$ & $29.57$  \\





\bottomrule
\end{tabular} }
\end{center}
   \caption{Human-Object Interaction Detection mAP on HICO-DET~\citep{hicodet}. Our method outperforms existing techniques over all splits of full, rare, and non-rare.}
\label{tab:sotahico}
\end{table}
\vspace{-3.5 mm}



\partitle{HICO-DET Results} are presented at Table~\ref{tab:sotahico}. Overall, Align-Former outperforms the other two techniques by $3.12$ mAP via ResNet-$50$ and $4.71$ mAP via ResNet-$101$ on all categories. This confirms that HO-I detection benefits from the end-to-end alignment of the targets and the predictions. Our improvement is even more pronounced on the rare split via $6.17$ mAP using ResNet-$101$, exhibiting the sample efficiency of our technique. 



\begin{table}[h]
\begin{center}
\resizebox{\columnwidth}{!}{%
\begin{tabular}{llcc|ccc}

\toprule 


Method   & Backbone & HICO-DET Pre-Trained? & Alignment-Supervised? & Agent & Scenario 1 & Scenario 2  \\ 
\midrule


\rowcolor{LightCyan}  Align-Former  & ResNet-$50$ & \xmark & \xmark & $24.63$ & $13.90$ & $14.15$  \\

\rowcolor{LightCyan}  Align-Former   & ResNet-$50$ & \cmark & \xmark 
& $\underline{27.95}$ & $\underline{15.52}$ & $\underline{16.06}$  \\

\rowcolor{LightCyan} Align-Former  & ResNet-$101$ & \xmark & \xmark 
& $20.00$ & $10.44$ & $10.79$  \\

\rowcolor{LightCyan} Align-Former  & ResNet-$101$ & \cmark & \xmark  & $\textbf{30.02}$ & $\textbf{15.82}$ & $\textbf{16.34}$ \\

  \cdashlinelr{1-7}

 Align-Former   & ResNet-$50$ & \xmark & \cmark & $66.78$ & $50.20$ & $56.42$  \\
 Align-Former   & ResNet-$101$ & \xmark & \cmark & $68.00$   & $55.40$ & $62.15$ \\ 





\bottomrule
\end{tabular} }
\end{center}
   \caption{Human-Object Interaction Detection mAP on V-COCO~\citep{vcoco}. Even though the performance is limited when trained from scratch on V-COCO, HICO-DET pre-training yields a considerable improvement on V-COCO.}
\label{tab:sotavcoco}
\end{table}

\vspace{-3.5 mm}

\partitle{V-COCO Results} are presented at Table~\ref{tab:sotavcoco}. We only compare to our own baselines~\footnote{Neither of the existing baselines (PPR-FCN and MX-HOI) evaluates on V-COCO. Additionally, strongly supervised stream of MX-HOI (No-Frills HO-I~\citep{gupta2018no}.) also is not evaluated on V-COCO}. We evaluate two different settings. \textit{i) Training on V-COCO from scratch}: Since the number of training images are quite limited (only $2k$ examples), training on V-COCO without alignment supervision yields limited accuracy on all three settings. \textit{ii) Transfer learning from HICO-Det}: where we fine-tune a HICO-DET pre-trained model on V-COCO. In all cases, pre-training on HICO-DET helps significantly. As one of the major goal of annotation-free training is the ability to pre-train on large-scale benchmarks, we see this as a promising direction in HO-I detection with cheap image-level supervision. 


We confirm that our model yields competitive performance on HICO-DET against competing benchmarks on all full, rare and non-rare splits, and showcases promising first results without alignment supervision on V-COCO, especially via transfer learning.

\begin{figure}[!tbp]
  \centering
  \begin{minipage}[t]{0.47\textwidth}
    \includegraphics[width=\textwidth]{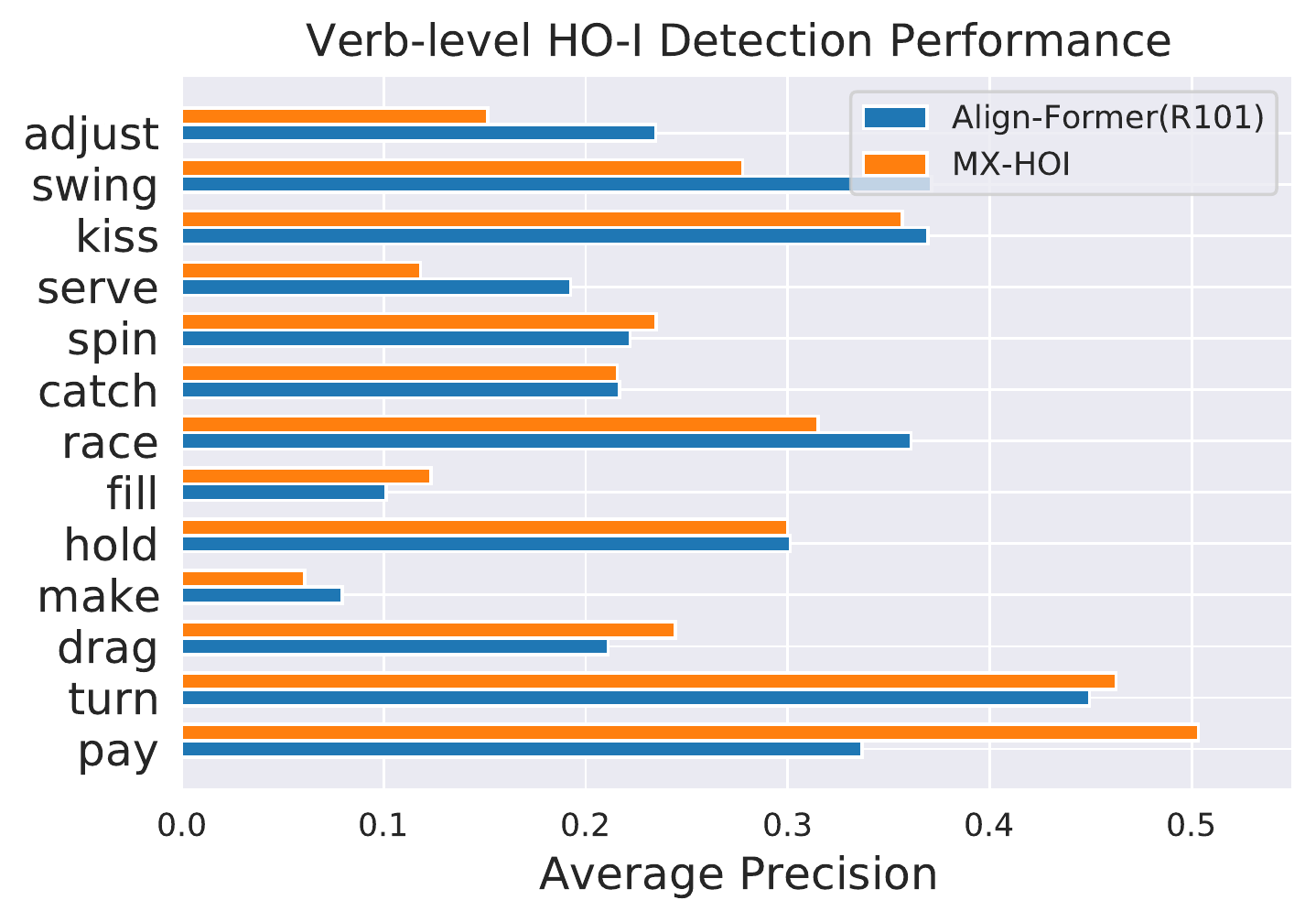}
\caption{Verb-level Performance on HICO-DET~\citep{hicodet} }
    \label{fig:class-ap}
  \end{minipage}
  \hfill
  \begin{minipage}[t]{0.47\textwidth}
    \includegraphics[width=\textwidth]{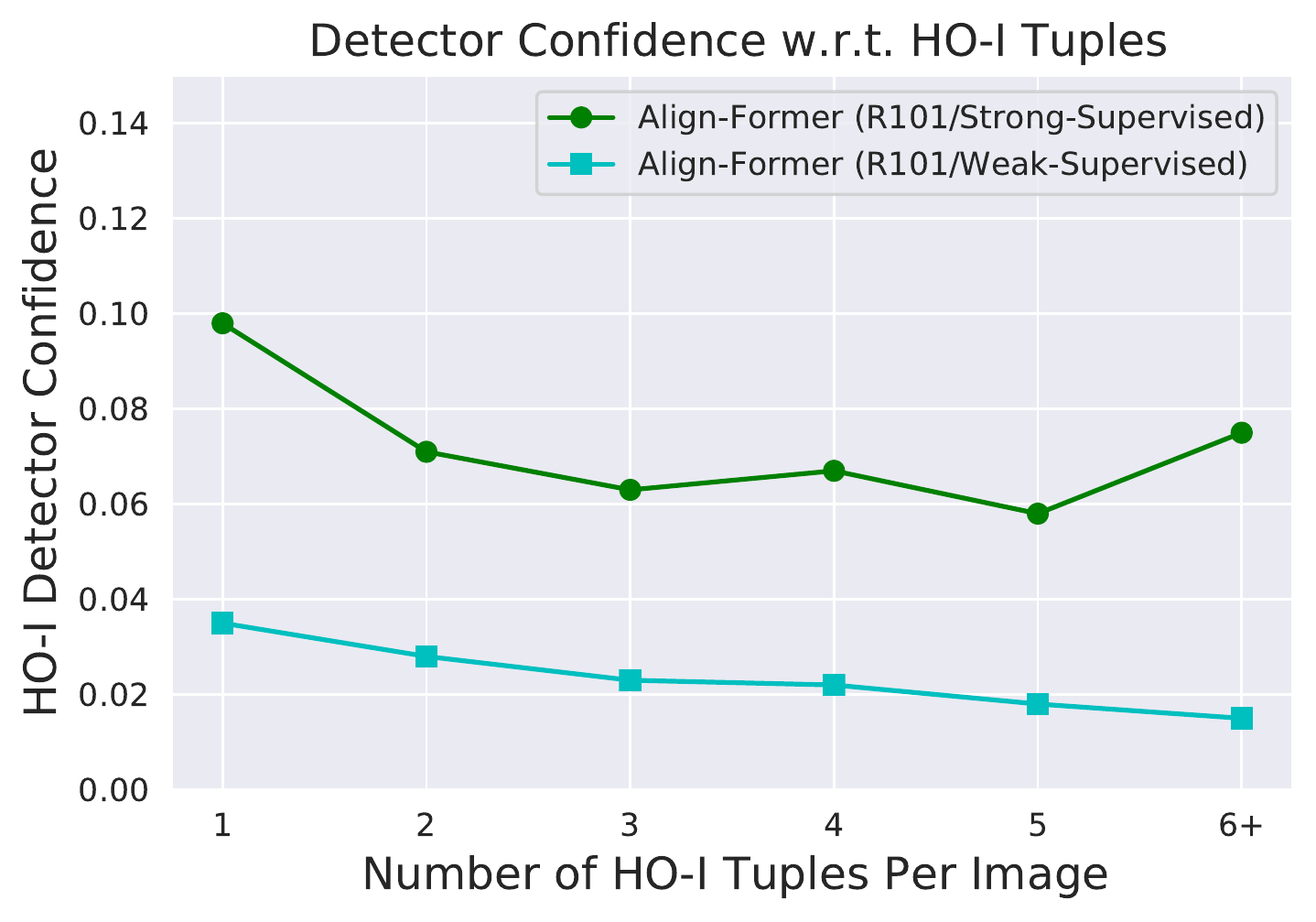}
    \caption{HO-I detector confidence w.r.t. number of HO-I tuples in an image on HICO-DET~\citep{hicodet}.}
        \label{fig:crowd}
  \end{minipage}

\end{figure}

\subsection{Further Analysis}






In this section, we provide analysis to better understand the contribution of Align-Former. 

\partitle{Verb-level Performance Comparison.} We visualize verb-level performance difference between weakly supervised Align-Former and MX-HOI in Figure~\ref{fig:class-ap}. We observe that Align-Former outperforms for pose and part-driven interactions like adjust, swing or kiss, while underperforming for scene-driven interactions like pay or turn. This indicates end-to-end learning of pose-based representations is more valuable than hand-crafted pose representations as in MX-HOI. For more results, refer to our Supp. material.




\vspace{-2 mm}
\partitle{W/ \textit{vs.} W/O Alignment Supervision.} To better understand the gap between strongly \textit{vs.} weakly supervised HO-I detection, we provide results of MX-HOI with strong supervision on HICO-DET in Table~\ref{tab:sotahico} as well as strongly supervised Align-Former in both datasets (Table~\ref{tab:sotahico}-~\ref{tab:sotavcoco}). Our method is flexible as it can be easily trained with strong and weak supervision with no change in architecture, whereas MX-HOI ensembles two CNNs (a weak~\citep{weaklyhoi1} and strong~\citep{gupta2018no} CNN) to do so. 


We have three main findings. \textit{i)} Weakly-supervised Align-Former outperforms strongly supervised MX-HOI on HICO-DET (Table~\ref{tab:sotahico}), which indicates our method compensates for the lack of supervision with its representational power. \textit{ii)} Strongly supervised Align-Former outperforms weakly supervised Align-Former on both datasets (Table~\ref{tab:sotahico}-~\ref{tab:sotavcoco}). This shows Align-Former better leverages the supervision when is used, and there is a room for improvement in weakly-supervised techniques. \textit{iii)} In Figure~\ref{fig:crowd}, we plot the confidence of strongly \textit{vs.} weakly supervised Align-Former as a function of number of HO-I tuples in an image on HICO-DET. As can be seen, strongly-supervised variant retains its performance whereas weakly-supervised degrades in confidence, which may help explain the performance gap between the two variants of Align-Former. 

\vspace{-2.5 mm}
\partitle{ResNet-101 vs. ResNet-50.} We implement Align-Former with ResNet-$50$ and $101$. Even though we do not observe significant difference at the verb- or object- level, the difference is at the interaction-level. Our findings are: \textit{i)} ResNet-$101$ outperforms $ResNet-50$ on both datasets across all settings, \textit{ii)} Surprisingly, ResNet-$101$ outperforms especially on the rare split of HICO-DET, and exhibits better transferability to V-COCO, despite higher number of parameters. 



 \begin{figure}[t]
    \centering
\includegraphics[width=0.56\textwidth]{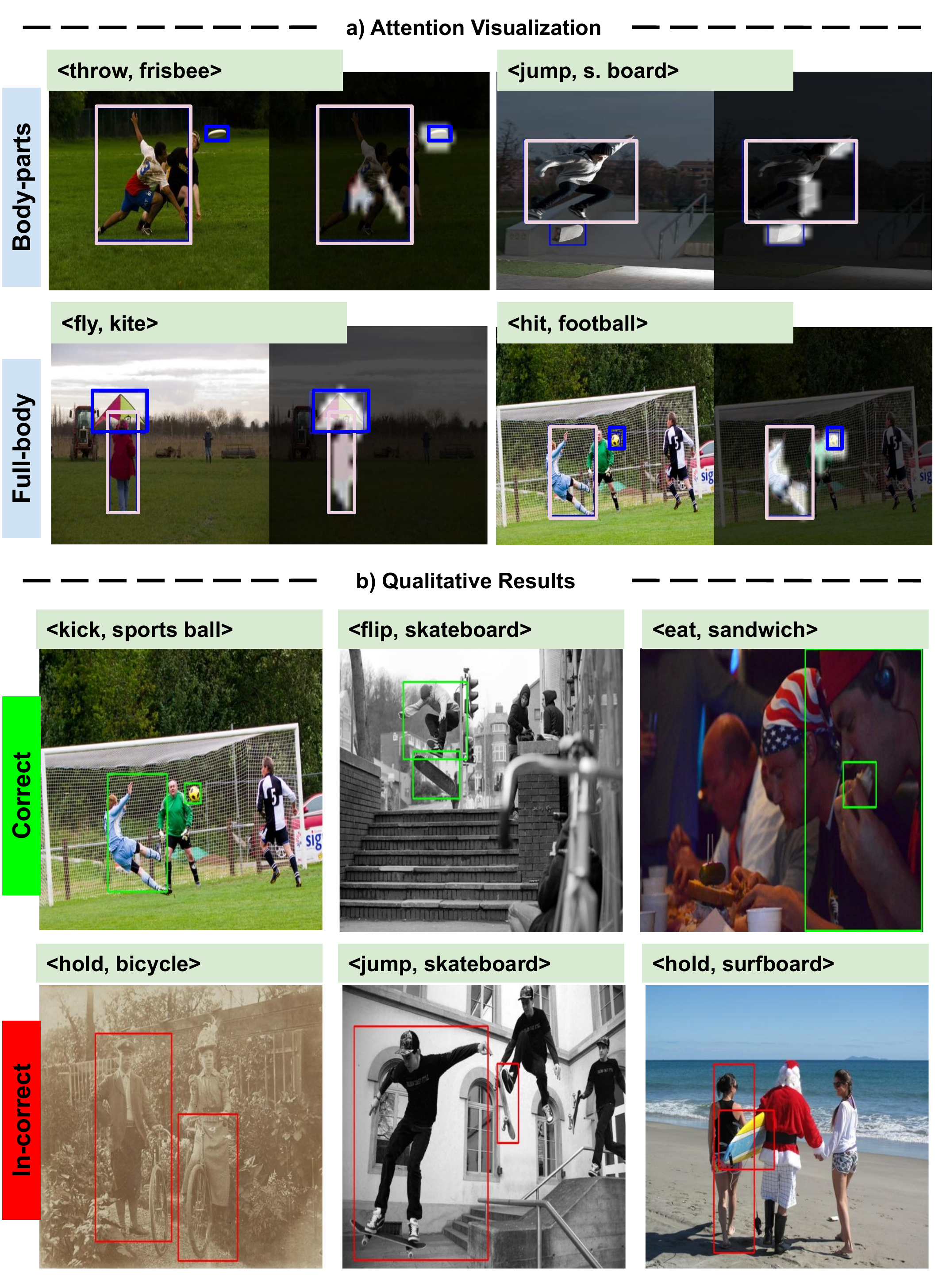}
 \caption{\textit{a)} Attention analysis of Align-Former reveals the focus on body-part and full-body. \textit{b)} Qualitative analysis of Align-Former reveals it can detect both dynamic and static interactions.}
    \label{fig:qualitative_joint}
\end{figure}

\vspace{-5 mm}



\partitle{Qualitative Inspection.} \textit{i) Attention Analysis}: To understand where Align-Former is looking at to perform HO-I alignment and detection, we present the attention matrix for a set of queries from the last layer of the Decoder in Figure~\ref{fig:qualitative_joint}-(\textcolor{red}{a}). We observe that Align-Former attends on body-parts when the visual information is sufficient, and full-body when the human-object has small scale. \textit{ii) Qualitative Results}: Finally, we visualize high-confident detection examples in Figure~\ref{fig:qualitative_joint}-(\textcolor{red}{b}). We observe that Align-Former can detect both dynamic interactions like \texttt{<kick, sports ball>} or static interactions like \texttt{<eat, sandwich>}. However, our method fails when humans can not be paired with their object of interaction, as is visualized in the bottom row. 

\vspace{-3 mm}

%% file: 5-conclusion.tex
\section{Conclusion}



This paper addressed HO-I detection from images. We proposed Align-Former, a visual-transformer based CNN that can learn to detect HO-I without alignment supervision, via image-level supervision. We equip Align-Former with HO-I align, a novel layer that learns to select correct detection targets based on geometric and visual priors. We show that Align-Former outperforms existing techniques for HO-I detection on HICO-DET especially on rare HO-I, and yields promising results on V-COCO, confirming the efficacy of our method. We hope our work inspires future research on reducing supervision in HO-I detection.

\partitle{Acknowledgements.} We are grateful for the constructive and useful feedback from all our anonymous reviewers as well as the meta reviewer, which helped us to greatly improve our manuscript.